\documentclass[10pt,twocolumn,letterpaper]{article}

\usepackage[pagenumbers]{cvpr} 

\usepackage{listings}
\usepackage{makecell}

\definecolor{cvprblue}{rgb}{0.21,0.49,0.74}
\usepackage[pagebackref,breaklinks,colorlinks,allcolors=cvprblue]{hyperref}

\def\paperID{*****} 
\def\confName{CVPR}
\def\confYear{2026}

\title{Can AI Understand Indonesian Children's Book Illustrations?
Compliance, Safety, and the Cultural Alignment Gap}

\author{Rian Adam Rajagede\\
Universitas Islam Indonesia\\
Sleman, Indonesia \\
{\tt\small rian.adam@uii.ac.id}
\and
Rochana Prih Hastuti\\
University of Central Florida\\
Orlando, FL, USA \\
{\tt\small rochana@ucf.edu}
\and
Fitri Hasanah Amhar\\
Grow the Seed\\
Indonesia \\
{\tt\small fitri@growtheseed.id}
}

\begin{document}

\maketitle

\begin{abstract}
\noindent
Children's book illustrations carry developmental, pedagogical, and cultural responsibilities that are often formally codified in guidelines. As AI systems begin entering children's publishing pipelines, both as evaluators and as generators, a natural question arises: can they reliably meet the structural, safety, and cultural standards required? We investigate this across three criteria: guideline compliance, safety, and cultural alignment, using Indonesia's Book Leveling Guideline as a government-issued evaluation framework with explicit visual criteria for four reading levels. To understand how well VLMs can identify and apply these guideline criteria, we first evaluate them on real published children's book pages and compare their assessments against a human baseline. To then examine whether AI-generated illustrations meet the same standards, we apply the VLM judge to images produced by two generation models and cross-check them against a subset of human illustrator raters. We find that VLMs are reliable on explicit criteria such as compliance and safety, but fall short on cultural alignment, systematically missing content that human raters flag as culturally inappropriate. AI-generated illustrations show the same blind spot: they satisfy structural and safety criteria but fail to capture implicit cultural details that human raters consistently identify.
\end{abstract}

\section{Introduction}

Children's book illustrations are not merely decorative. They carry developmental, pedagogical, and cultural responsibilities: shaping early literacy, representing the child's world, and reflecting the social norms of the communities that produce them~\cite{KellyWare2019USINGPI,Zhang2023HowDP}. In many countries, these responsibilities are formally codified. Indonesia's \textit{Pedoman Perjenjangan Buku} (Book Leveling Guideline), issued by The Ministry of Primary and Secondary Education, specifies visual criteria for illustrations across four reading levels (pre-readers to early-chapter readers), covering layout dominance, scene complexity, and dialogue structure~\cite{ghozalli2020panduan}.

At the same time, AI systems are disrupting creative industry workflows, including the children's book creation process. Images generated by text-to-image models~\cite{stable_diffusion,flux,imagen} are beginning to appear in illustrated books. Recent work has documented this shift at the family level: Jin and Yuan~\cite{jin2025they} found that children aged 4--8 rarely noticed or cared whether story images were AI-generated or human-created, and that parents generally accepted AI-generated illustrations provided they met perceived quality thresholds around visual consistency and emotional clarity. Yet that same study found safety to be a binary, non-negotiable threshold for parents, with content that fails it rejected outright, regardless of other qualities. As AI-generated illustrations enter publishing pipelines for children, whether they meet domain-specific quality and safety standards is therefore not just an academic question. This question is not trivial to answer, as most evaluations of text-to-image models rely on generic perceptual metrics, such as FID or CLIPScore~\cite{jayasumana2024rethinking} or aesthetics and plausibility~\cite{liang2024rich}.

\begin{figure}[tbp]
    \centering
    \begin{minipage}{0.155\textwidth}
        \centering
        \includegraphics[width=\textwidth]{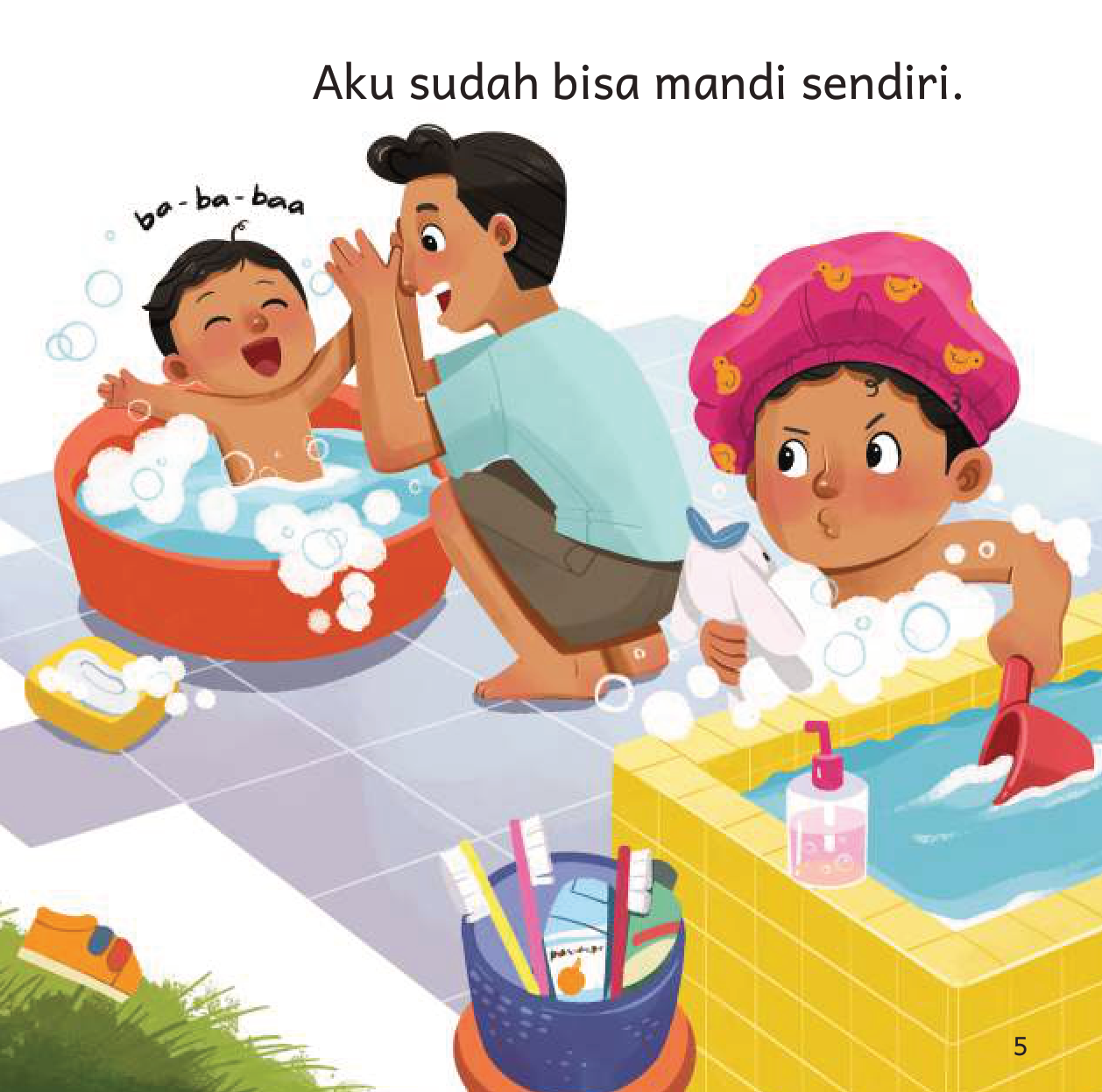}
    \end{minipage}
    \hfill 
    \begin{minipage}{0.155\textwidth}
        \centering
        \includegraphics[width=\textwidth]{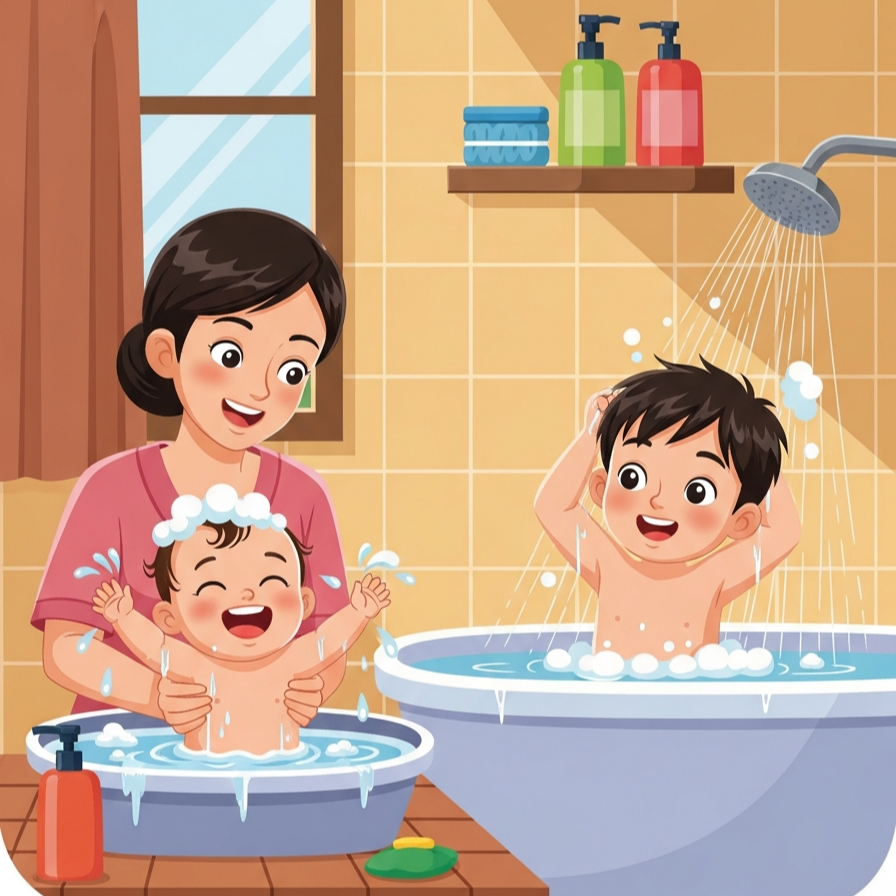}
    \end{minipage}
    \hfill 
    \begin{minipage}{0.155\textwidth}
        \centering
        \includegraphics[width=\textwidth]{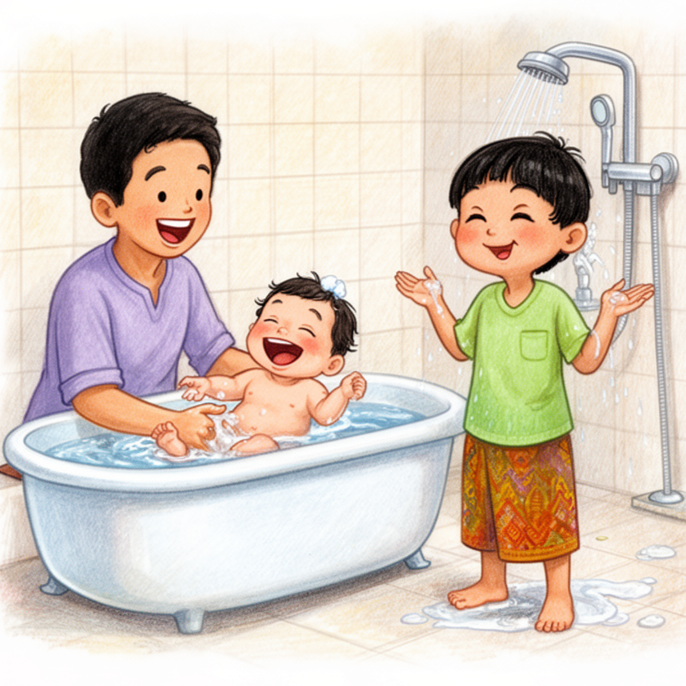}
    \end{minipage}

    \caption{Left: original human illustration. Middle and right: AI-generated images by Imagen 4 and FLUX.1 [dev], respectively. Both AI images depict a shower fixture, which is uncommon in Indonesian homes where bathing typically uses a dipper from a standing tub. This cultural detail is captured by the human illustrator but missed by both generators.}
    \label{fig:side_by_side}
\end{figure}

We use Indonesian children's books as a case study for two reasons. First, the Book Leveling Guideline provides a concrete, written evaluation anchor. Its structural rules, covering illustration dominance to typography usage per reading level, are explicit and verifiable. This makes it possible to measure how well VLMs and generative models comply with the guideline without relying solely on subjective impressions. Second, Indonesia's rich and distinct culture~\cite{aji2022one} introduces a genuine challenge for children's book illustration. As we can see in Figure~\ref{fig:side_by_side}, which shows two illustrations side by side: the left illustration was created by a human, and the right was generated by AI. While, at a glance, the AI illustration quality is on par, the AI missed a small cultural detail: it is not common to use showers in Indonesia. Capturing cultural specificity is a challenging task in AI research. Prior works have shown that systems trained predominantly on Western-sourced data underrepresent the cultural knowledge of other regions~\cite{ng2025sea, pranida-etal-2025-culturally, cahyawijaya2025crowdsource}. Cultural content rules are not written in the guidelines. What an Indonesian home, child, or outfit should look like exists as implicit knowledge, known by illustrators from their background and training, and for illustrations targeting Indonesian children, getting this right is far from trivial. 

This motivates two research questions. \textbf{RQ1:} How well do vision-language models (VLMs) align with human expert raters when evaluating real, published Indonesian children's book illustrations against the formal guideline criteria? \textbf{RQ2:} Where do text-to-image models comply with, and where do they diverge from, those criteria, particularly on dimensions the guideline leaves implicit, such as cultural representation and safety? To answer these questions, we conduct a two-task study. In Task~1, we measure VLM--human alignment on published book pages rated against the Guideline. In Task~2, we extend this evaluation to illustrations produced by two text-to-image models, using VLM scores cross-checked against human illustrator ratings on a subset. Both tasks share a common three-pillar framework: guideline compliance, cultural alignment, and safety, ordered from most explicit to most implicit in the guideline.


\section{Study Design}

\subsection{Book Leveling Guideline and Dataset}

\begin{figure}[tbp]
\centering
\includegraphics[width=0.43\textwidth]{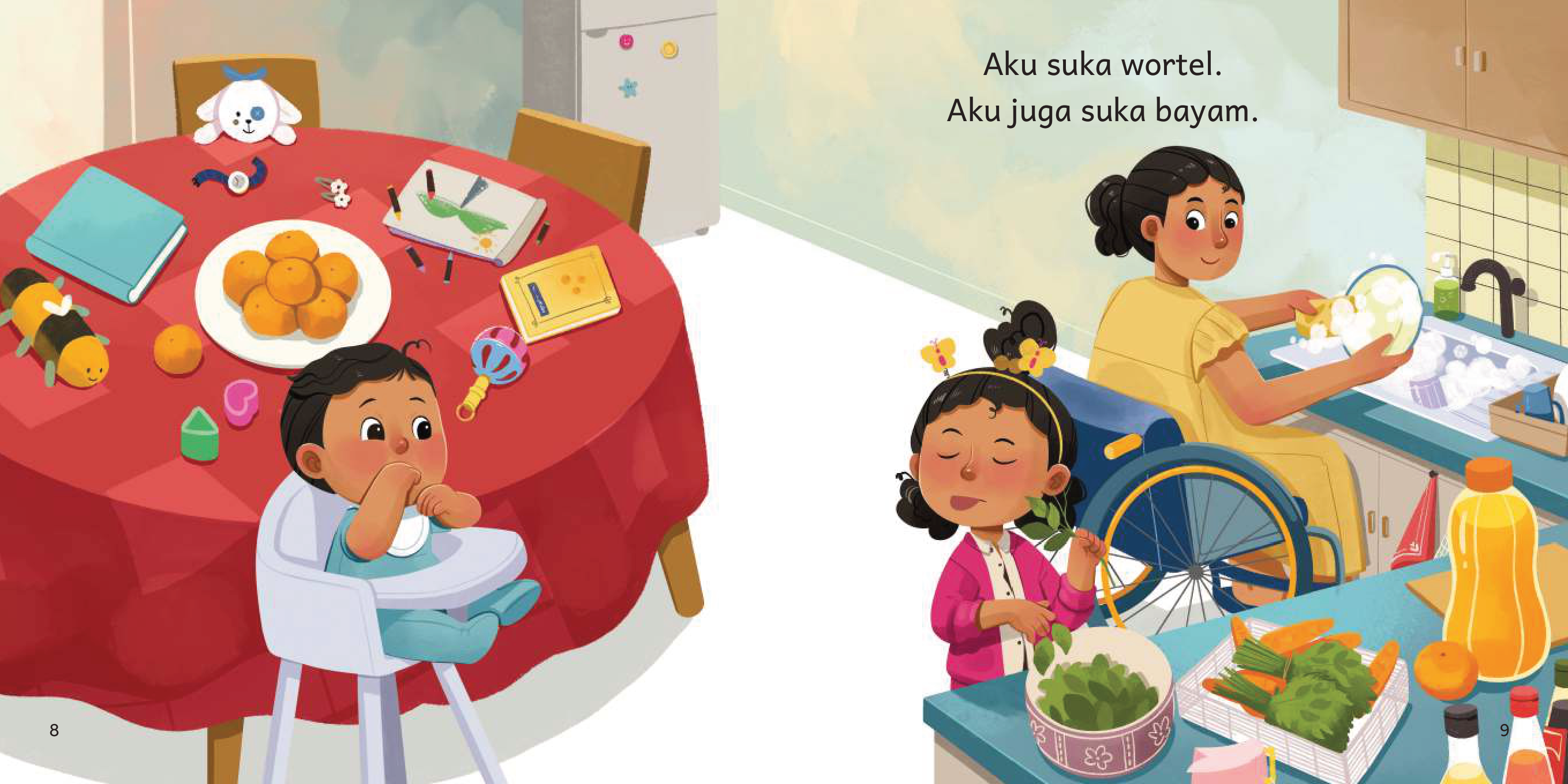}
\caption{A sample page from reading level A (0--7 yr). The guideline checklist for this level requires: 2D/3D illustration style, illustration dominant over text, full color, no dialogue balloons, and at most 3 sentences.}
\label{fig:jenjang_examples}
\end{figure}

Indonesia's Book Leveling Guideline \cite{ghozalli2020panduan} classifies books by reading level from emergent to advanced. Table~\ref{tab:jenjang_classification} shows the four reader classifications used in this study, and Figure~\ref{fig:jenjang_examples} shows representative pages alongside the key illustration criteria. Our study covers the four early-literacy levels relevant to picture books. Book images dataset is sourced from the Ministry's digital library at \url{https://buku.kemendikdasmen.go.id}.

\begin{table}[h]
\centering
\caption{Reader classification in Indonesia's Book Leveling Guideline.}
\label{tab:jenjang_classification}
\resizebox{\columnwidth}{!}{%
\begin{tabular}{lllp{4.5cm}}
\toprule
\textbf{Level} & \textbf{Reader Profile} & \textbf{Ages} & \textbf{Reading Ability} \\
\midrule
A    & Emergent Reader       & 0--7   & Emergent; requires full scaffolding \\
B1   & Early/Beginning Reader & 6--8   & Short sentences; builds reading accuracy \\
B2   & Early/Beginning Reader & 7--9   & Moderate sentences; richer vocabulary \\
B3   & Early/Beginning Reader & 8--10  & Simple paragraphs; dialogue introduced \\
\bottomrule
\end{tabular}
}
\end{table}

\subsection{Evaluation Framework}

We assess each page across three dimensions: \textbf{Guideline Compliance} measures structural rule-following (e.g., balloon presence) that can be verified without cultural knowledge, \textbf{Cultural Alignment} measures whether Indonesian-specific elements (e.g., clothing) are present and accurate, \textbf{Safety} measures freedom from age-inappropriate or harmful content.

\subsection{Task 1: Alignment on Published Pages}
\label{sec:task1}

Task 1 measures how well VLMs align with human raters when evaluating real, published Indonesian children's book pages against the guidelines. We sample 100 pages from the dataset (25 per level, stratified across titles) and evaluate them with three VLMs: Claude Sonnet 4.6 (Commercial LLM), Qwen3-VL 8B~\cite{qwen3technicalreport} (open-weight general model), and SEA-LION-v4-8B-VL~\cite{ng2025sea} (open-weight Qwen fine-tuned on South East Asian cultural data). Each model receives the page image, the page text, and an instruction requesting level prediction, guideline feature extraction, cultural alignment score, a safety flag, and overall age appropriateness, with the full guideline appended as reference.

For the human baseline, three professional annotators independently evaluated a random sample of 20 pages drawn from the 100-page pool (5 per level), rating level classification, cultural alignment, and safety.


\subsection{Task 2: Image Generation and Evaluation}
\label{sec:task2}

Task 2 extends the measurement to AI-generated illustrations. We select 8 books (2 per level) and generate 3 pages per book using two text-to-image models: Imagen 4~\cite{imagen} and FLUX.1 [dev]~\cite{flux}, yielding 24 images per generator (48 total). Generation prompts are constructed using a round-trip strategy: Claude Sonnet 4.6 first produces a structured visual description by reading the original published page image; then, the page text and level illustration requirements are combined with explicit guidance on the Indonesian cultural context to form the final prompt. This approach separates language understanding from image generation, producing richer, more guideline-aware prompts than direct page-to-image generation.

The 48 generated images are then evaluated by Claude Sonnet 4.6 as a judge using a similar Task~1-structured prompt, enabling direct comparison of VLM scores across real and AI-generated pages. To cross-check VLM scores on a subset, three professional illustrators evaluated 12 generated page-pairs from level A (2 books, 3 pages each, both generators). Raters scored each image on four criteria (visual guidelines, cultural elements, age appropriateness, and logical coherence) on a 1--3 scale.

\section{Results and Discussion}

\subsection{Task 1: Alignment on Published Pages }

\begin{figure}[h]
\centering
\includegraphics[width=0.85\columnwidth]{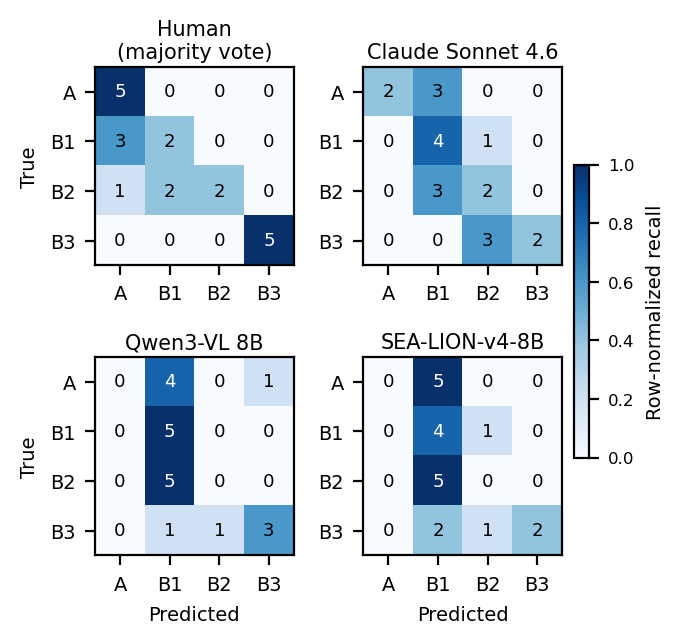}
\caption{Row-normalized confusion matrices for level classification (N=20). Cell values are raw counts. A and B3 are anchor levels; B2 is the hardest level for all annotators.}
\label{fig:confusion}
\end{figure}

\textbf{Book level prediction.} Addressing RQ1, Figure~\ref{fig:confusion} shows confusion matrices for the N=20 human-annotated subset. Qwen and SEA-LION effectively collapse the four-level scale to two classes, correctly separating B1 while folding levels A and B2 into it. SEA-LION, a fine-tuned variant of Qwen with South East Asian cultural data, reproduces this failure identically, suggesting the fine-tuning did not introduce new visual discrimination capacity. Claude identifies level A more reliably but still collapses B2 into B1. Human raters perfectly separate A and B3 but show notable B1/B2 confusion, indicating this boundary is genuinely ambiguous even to experts. On off-by-one (OB1) accuracy, where a prediction is correct if it falls within one level of the true label, all VLMs approach human performance; the gap is concentrated in exact level separation, particularly at B2.

\textbf{Cultural Alignment and Safety.} Table~\ref{tab:cultural_safety} shows results on the N=20 human subset. Safety false positives are zero for all VLMs. Cultural alignment averages are close across annotators, but the top-rating distribution reveals a systematic divergence: human raters assign the highest cultural rating in 23.3\% of cases; Claude in 15.0\%, and both Qwen and SEA-LION in 0.0\%. \textbf{VLMs can detect cultural absence but fail to recognize cultural richness}, even when human raters converge on clear cultural presence.

\begin{table}[h]
\centering
\caption{Cultural alignment and safety on the 20-page human subset. Cultural score is on a 1--3 scale. Top-rating \% is the share of pages receiving the highest cultural rating. Safety FP = rate of incorrectly flagging safe pages.}
\label{tab:cultural_safety}
\resizebox{\columnwidth}{!}{%
\begin{tabular}{lccc}
\toprule
\textbf{Model} & \textbf{Cult.\ Avg} & \textbf{Top-rating \%} & \textbf{Safety FP} \\
\midrule
Human (majority) & 1.92 & 23.3\% & 0.0\% \\
Claude Sonnet 4.6 & 1.85 & 15.0\% & 0.0\% \\
Qwen3-VL 8B      & 1.60 & 0.0\%  & 0.0\% \\
SEA-LION-v4-8B   & 1.85 & 0.0\%  & 0.0\% \\
\bottomrule
\end{tabular}
}
\end{table}

\begin{figure}[h]
\centering
\includegraphics[width=0.4\textwidth]{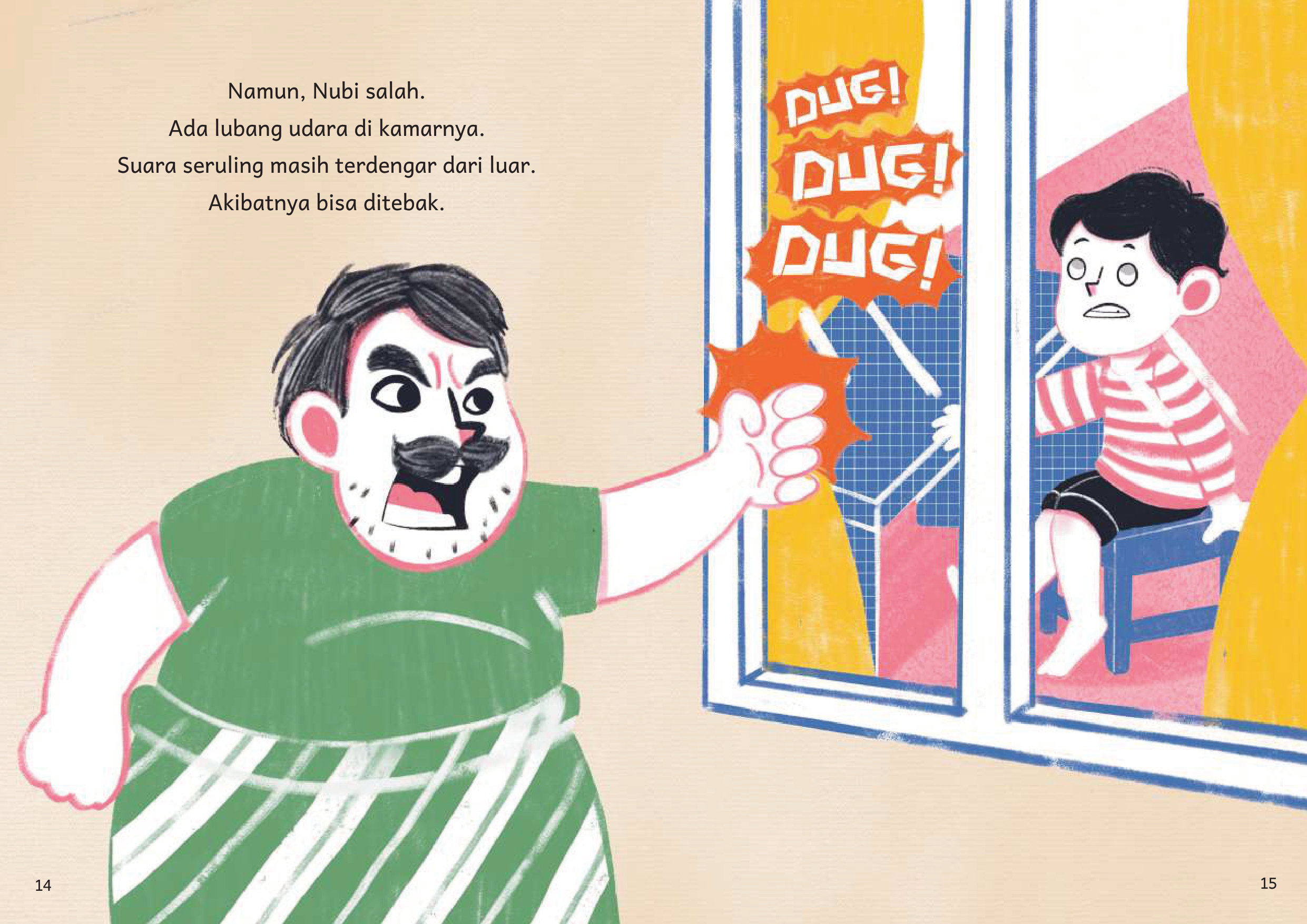}
\caption{The one page from Level B2 flagged as potentially unsafe by Claude.}
\label{fig:safety_flag}
\end{figure}

On the full N=100 evaluation, Claude flagged one safety concern (shown in Figure~\ref{fig:safety_flag}), depicting an adult aggressively knocking on a window toward a frightened child, which could be distressing for young readers. Although this is a false positive on a published book, it demonstrates that Claude has safety sensitivity.

\subsection{Task 2: AI-Generated Illustration Evaluation}

Addressing RQ2, Table~\ref{tab:task2_human} reports human evaluator scores for the two generators on Level A. Imagen 4 outperforms FLUX.1 [dev] on every criterion; the largest gap is on visual guideline adherence. Both models score below the midpoint on cultural elements. In comparative preference responses, the most frequent outcome was ``Both Bad'' (42.2\%), indicating that neither model meets professional illustrator standards.

\begin{table}[h]
\centering
\caption{Task 2 human evaluator scores per criterion (scale 1--3, higher is better). Level A, 3 raters, 2 books, 3 pages each.}
\label{tab:task2_human}
\resizebox{\columnwidth}{!}{%
\begin{tabular}{lcc}
\toprule
\textbf{Criterion} & \textbf{FLUX.1 [dev]} & \textbf{Imagen 4} \\
\midrule
Q1: Visual guidelines     & 1.35 & 2.18 \\
Q2: Cultural elements     & 1.53 & 1.76 \\
Q3: Age-appropriateness   & 1.65 & 2.12 \\
Q4: Logical coherence     & 1.59 & 1.87 \\
\midrule
\textbf{Overall}          & \textbf{1.53} & \textbf{1.97} \\
\midrule
Preferred (\%)            & 12.5 & 40.6 \\
Both Good (\%)             & \multicolumn{2}{c}{4.7} \\
Both Bad (\%)             & \multicolumn{2}{c}{42.2} \\
\bottomrule
\end{tabular}
}
\end{table}


\textbf{VLM evaluation on all 48 images.} Across all 48 generated pages, VLM scores are consistently higher than human scores on the overlapping Level~A subset, but the relative model rankings largely agree: Imagen scores above FLUX on visual guidelines (2.54 vs.\ 2.33) and logical coherence (2.46 vs.\ 2.38), and both score equally on age-appropriateness (2.79), all matching human raters. The one exception is cultural elements, where the VLM ranks FLUX above Imagen (2.25 vs.\ 2.08) while humans rate it the opposite way, repeating the same cultural gap observed in Task~1 where the VLM incorrectly rates cultural elements as present when human raters, who carry implicit cultural knowledge, judge them as lacking.

\subsection{Qualitative Limits: Expressiveness, Gesture, and the Cultural Gap}

\begin{figure}[tbp]
    \centering
    \begin{minipage}{0.155\textwidth}
        \centering
        \includegraphics[width=\textwidth]{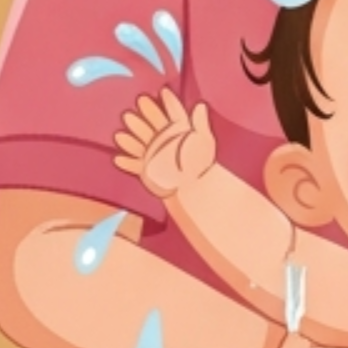}
    \end{minipage}
    \hfill 
    \begin{minipage}{0.155\textwidth}
        \centering
        \includegraphics[width=\textwidth]{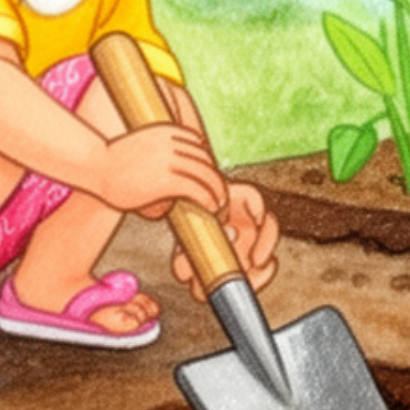}
    \end{minipage}
    \hfill
    \begin{minipage}{0.155\textwidth}
        \centering
        \includegraphics[width=\textwidth]{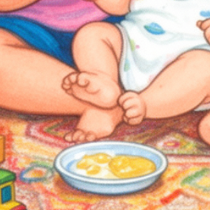}
    \end{minipage}

    \caption{Examples of anatomy failures in AI-generated pages. Left: Imagen 4. Middle and right: FLUX.1 [dev].}
    \label{fig:gesture_anatomy}
\end{figure}

\paragraph{Expressiveness and gesture as an unwritten criterion.} The open-ended feedback from human illustrators in Task 2 reveals a dimension that neither the guideline nor the structured rubric captures: expressive coherence. FLUX pages were repeatedly flagged for emotionally unconvincing characters, such as "strange child emotions" or "not engaging through gesture", concerns that are orthogonal to guideline compliance, cultural elements, or age-appropriateness as defined. Anatomy failures (hands, feet, limb count) appeared across both generators, as shown in Figure~\ref{fig:gesture_anatomy}. These concerns point to another evaluation dimension beyond the current framework: whether illustrated characters feel emotionally and physically plausible.

\begin{figure}[h]
    \centering
    \includegraphics[width=0.35\textwidth]{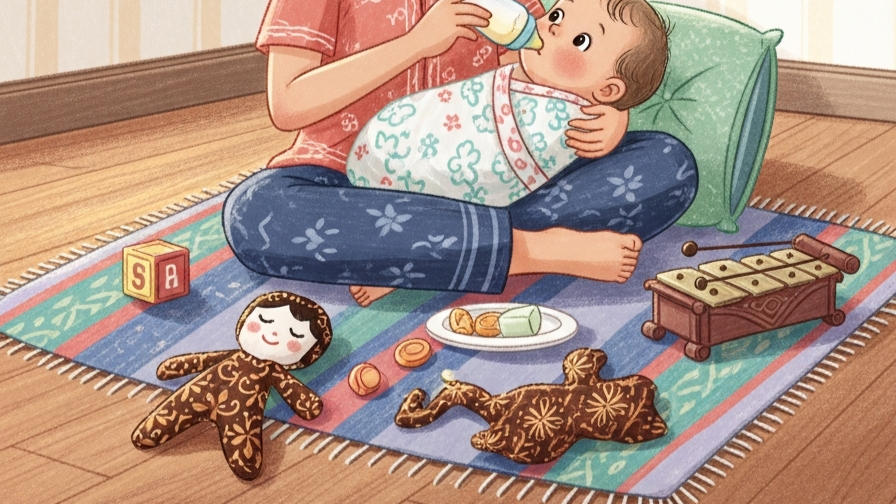}
    \caption{An Imagen 4 generated page where batik patterns appear on the mat, clothing, and toys with similar repetitions. While batik is a real Indonesian cultural element, the uniform overuse across all objects was flagged by an annotator as unrealistic.}
    \label{fig:batik_overuse}
\end{figure}

\paragraph{Cultural absence is detectable; cultural richness is not.} Human illustrators showed high inter-rater agreement on FLUX's cultural element scores without any explicit rubric, converging on identical cultural ratings far more often than on visual guidelines or age-appropriateness. Cultural absence was categorical, not a matter of interpretation. A clear example is the shower scene (Figure~\ref{fig:side_by_side}): both models depict a bathroom with a shower fixture, while in Indonesian homes bathing typically uses a dipper from a standing tub. A contrasting case appears in Figure~\ref{fig:batik_overuse}, where Imagen 4 places batik patterns on the mat, clothing, and toys all at once with nearly identical repetitions. While batik is a genuine Indonesian cultural marker, applying it uniformly across every surface in a scene is not how it appears in everyday life. One annotator flagged this as visually unrealistic, pointing to a cultural overemphasis in which the model adds recognizable cultural signals without the implicit knowledge of how and where they actually appear. This reveals an asymmetric structure: cultural \textit{absence} is reliably detectable, while recognizing cultural \textit{richness} requires implicit knowledge that current models do not have.

\section{Conclusion}

Both VLM evaluators and generative models handle explicit, written rules well: level-prediction approaches match human OB1 accuracy, and safety flags align with human judgment. Implicit cultural knowledge is a different problem. VLMs can detect cultural absence but fail to recognize cultural richness, and neither Imagen 4 nor FLUX.1 [dev] meets the implicit cultural expectations a professional illustrator would apply.


{
    \small
    \bibliographystyle{ieeenat_fullname}
    \bibliography{main}

@article{Zhang2023HowDP,
  title={How do Preschool Teachers Teach Cultural Diversity Using Picture Books?},
  author={Xiaoyue Zhang},
  journal={Lecture Notes in Education Psychology and Public Media},
  year={2023},
  url={https://api.semanticscholar.org/CorpusID:258156817}
}

@inproceedings{cahyawijaya2025crowdsource,
  title={Crowdsource, crawl, or generate? creating sea-vl, a multicultural vision-language dataset for southeast asia},
  author={Cahyawijaya, Samuel and Lovenia, Holy and Moniz, Joel Ruben Antony and Wong, Tack Hwa and Farhansyah, Mohammad Rifqi and Maung, Thant Thiri and Hudi, Frederikus and Anugraha, David and Habibi, Muhammad Ravi Shulthan and Qorib, Muhammad Reza and others},
  booktitle={Proceedings of the 63rd Annual Meeting of the Association for Computational Linguistics (Volume 1: Long Papers)},
  pages={18685--18717},
  year={2025}
}

@inproceedings{pranida-etal-2025-culturally,
    title = "Culturally-Nuanced Story Generation for Reasoning in Low-Resource Languages: The Case of {J}avanese and {S}undanese",
    author = "Pranida, Salsabila Zahirah  and
      Genadi, Rifo Ahmad  and
      Koto, Fajri",
    editor = "Adelani, David Ifeoluwa  and
      Arnett, Catherine  and
      Ataman, Duygu  and
      Chang, Tyler A.  and
      Gonen, Hila  and
      Raja, Rahul  and
      Schmidt, Fabian  and
      Stap, David  and
      Wang, Jiayi",
    booktitle = "Proceedings of the 5th Workshop on Multilingual Representation Learning (MRL 2025)",
    month = nov,
    year = "2025",
    address = "Suzhuo, China",
    publisher = "Association for Computational Linguistics",
    url = "https://aclanthology.org/2025.mrl-main.25/",
    doi = "10.18653/v1/2025.mrl-main.25",
    pages = "369--384",
    ISBN = "979-8-89176-345-6"
}

@misc{qwen3technicalreport,
      title={Qwen3 Technical Report}, 
      author={Qwen Team},
      year={2025},
      eprint={2505.09388},
      archivePrefix={arXiv},
      primaryClass={cs.CL},
      url={https://arxiv.org/abs/2505.09388}, 
}

@article{ghozalli2020panduan,
  title={Panduan mengilustrasi dan mendesain cerita anak untuk tenaga profesional},
  author={Ghozalli, Evelyn},
  journal={Jakarta Timur: Pusat Pembinaan Bahasa dan Sastra, Badan Pengembangan dan Pembinaan Bahasa, Kementerian Pendidikan dan Kebudayaan Republik Indonesia},
  year={2020}
}

@inproceedings{ng2025sea,
  title={Sea-lion: Southeast asian languages in one network},
  author={Ng, Raymond and Nguyen, Thanh Ngan and Yuli, Huang and Chia, Tai Ngee and Yi, Leong Wai and Leong, Wei Qi and Yong, Xianbin and Ngui, Jian Gang and Susanto, Yosephine and Cheng, Nicholas and others},
  booktitle={Proceedings of the 14th International Joint Conference on Natural Language Processing and the 4th Conference of the Asia-Pacific Chapter of the Association for Computational Linguistics},
  pages={512--526},
  year={2025}
}

@inproceedings{KellyWare2019USINGPI,
  title={USING PICTUREBOOK ILLUSTRATIONS TO HELP YOUNG CHILDREN UNDERSTAND DIVERSITY},
  author={Janette Kelly-Ware and Nicola Daly},
  year={2019},
  url={https://api.semanticscholar.org/CorpusID:235175872}
}

@inproceedings{aji2022one,
  title={One country, 700+ languages: NLP challenges for underrepresented languages and dialects in Indonesia},
  author={Aji, Alham Fikri and Winata, Genta Indra and Koto, Fajri and Cahyawijaya, Samuel and Romadhony, Ade and Mahendra, Rahmad and Kurniawan, Kemal and Moeljadi, David and Prasojo, Radityo Eko and Baldwin, Timothy and others},
  booktitle={Proceedings of the 60th Annual Meeting of the Association for Computational Linguistics (Volume 1: Long Papers)},
  pages={7226--7249},
  year={2022}
}

@inproceedings{liang2024rich,
  title={Rich human feedback for text-to-image generation},
  author={Liang, Youwei and He, Junfeng and Li, Gang and Li, Peizhao and Klimovskiy, Arseniy and Carolan, Nicholas and Sun, Jiao and Pont-Tuset, Jordi and Young, Sarah and Yang, Feng and others},
  booktitle={Proceedings of the IEEE/CVF Conference on Computer Vision and Pattern Recognition},
  pages={19401--19411},
  year={2024}
}

@inproceedings{jayasumana2024rethinking,
  title={Rethinking fid: Towards a better evaluation metric for image generation},
  author={Jayasumana, Sadeep and Ramalingam, Srikumar and Veit, Andreas and Glasner, Daniel and Chakrabarti, Ayan and Kumar, Sanjiv},
  booktitle={Proceedings of the IEEE/CVF conference on computer vision and pattern recognition},
  pages={9307--9315},
  year={2024}
}

@article{jin2025they,
  title={“They all look mad with each other”: Understanding the needs and preferences of children and parents in AI-generated images for stories},
  author={Jin, Qiao and Yuan, Ye},
  journal={International Journal of Child-Computer Interaction},
  pages={100787},
  year={2025},
  publisher={Elsevier}
}

@inproceedings{stable_diffusion,
  title={Scaling rectified flow transformers for high-resolution image synthesis},
  author={Esser, Patrick and Kulal, Sumith and Blattmann, Andreas and Entezari, Rahim and M{\"u}ller, Jonas and Saini, Harry and Levi, Yam and Lorenz, Dominik and Sauer, Axel and Boesel, Frederic and others},
  booktitle={Forty-first international conference on machine learning},
  year={2024}
}

@misc{imagen,
  author       = {{Google DeepMind}},
  title        = {Imagen: Our Leading Text-to-Image Model},
  year         = {2025},
  howpublished = {\url{https://deepmind.google/models/imagen/}},
  note         = {Accessed: 2026-04-12}
}

@misc{flux,
      title={FLUX.1 Kontext: Flow Matching for In-Context Image Generation and Editing in Latent Space},
      author={Black Forest Labs and Stephen Batifol and Andreas Blattmann and Frederic Boesel and Saksham Consul and Cyril Diagne and Tim Dockhorn and Jack English and Zion English and Patrick Esser and Sumith Kulal and Kyle Lacey and Yam Levi and Cheng Li and Dominik Lorenz and Jonas Müller and Dustin Podell and Robin Rombach and Harry Saini and Axel Sauer and Luke Smith},
      year={2025},
      eprint={2506.15742},
      archivePrefix={arXiv},
      primaryClass={cs.GR},
      url={https://arxiv.org/abs/2506.15742},
}
}

\end{document}